\documentclass[letterpaper, 10 pt, conference]{ieeeconf}  % Comment this line out if you need a4paper

\IEEEoverridecommandlockouts                              % This command is only needed if 
\usepackage{graphics} % for pdf, bitmapped graphics files
\usepackage{epsfig} % for postscript graphics files
\usepackage{times} % assumes new font selection scheme installed
\usepackage{amsmath} % assumes amsmath package installed
\usepackage{amssymb}  % assumes amsmath package installed
\usepackage{float}
\usepackage[caption = false]{subfig}
\usepackage{afterpage}
\usepackage[table]{xcolor}
\usepackage{booktabs}
\usepackage{caption}
\usepackage{amsfonts}
\usepackage{algorithm,algpseudocode}

\usepackage{balance}
\usepackage{dblfloatfix}
\title{\LARGE \bf
Bi-Level Image-Guided Ergodic Exploration with Applications to Planetary Rovers
}

\author{Elena Wittemyer$^{*\dagger}$ and Ian Abraham$^\dagger$% <-this % stops a space
\thanks{$^*$This work was supported by Yale College First-Year Summer Research Fellowship in the Sciences and Engineering. }% <-this % stops a space
\thanks{$^\dagger$All authors are in the Department of Mechanical Engineering,
        Yale University, New Haven, CT USA.}\\
\thanks{\tt\small Email: \{ian.abraham, elena.wittermyer\} @yale.edu}%
}

\begin{document}

\maketitle
\thispagestyle{empty}
\pagestyle{empty}

%%%%%%%%%%%%%%%%%%%%%%%%%%%%%%%%%%%%%%%%%%%%%%%%%%%%%%%%%%%%%%%%%%%%%%%%%%%%%%%%
\begin{abstract}
    We present a method for image-guided exploration for mobile robotic systems. Our approach extends ergodic exploration methods, a recent exploration approach that prioritizes complete coverage of a space, with the use of a learned image classifier that automatically detects objects and updates an information map to guide further exploration and localization of objects. Additionally, to improve outcomes of the information collected by our robot's visual sensor, we present a decomposition of the ergodic optimization problem as bi-level \emph{coarse} and \emph{fine} solvers, which act respectively on the robot's body and the robot's visual sensor.
    
    %However, to consider both of these entities in a single trajectory optimization is computationally infeasible. Rather, we propose a method of hierarchically decomposing the trajectory optimization into two, linearly-performed optimizations that yield favorable visual sensor information while maintaining reasonable computational effort. 
    Our approach is applied to geological survey and localization of rock formations for Mars rovers, with real images from Mars rovers used to train the image classifier. 
    Results demonstrate 1) improved localization of rock formations compared to naive approaches while 2) minimizing the path length of the exploration through the bi-level exploration.
    %improved speed performance of generating ergodic exploratory trajectories with minimal loss of performance. 
    %using the decomposed trajectory approach (a) achieves improved coverage in short run times, and (b) outperforms one approach in which the sensor is kept fixed and a second approach in which a trajectory is generated randomly in target localization, despite these approaches requiring less computational effort and achieving greater coverage.
\end{abstract}

%%%%%%%%%%%%%%%%%%%%%%%%%%%%%%%%%%%%%%%%%%%%%%%%%%%%%%%%%%%%%%%%%%%%%%%%%%%%%%%%
\section{INTRODUCTION}

    In many extreme environments, effective autonomy in mobile robots is critical to the expedited success of a mission. For Mars rovers, routes are planned by engineers in an involved process made time-intensive by the 5-20 minute communication delay \cite{ormston_2012} between Earth and Mars. During the communications delay, NASA scientists implement autonomous navigation software that allows rovers to identify simple hazards over short expanses of terrain to circumvent these obstacles. However, the regions and features of interest a rover visits are still planned by scientists on Earth using images received from rovers of nearby terrain. To enable rovers to explore larger regions of minimally-charted terrain, local, sensor-guided trajectory planning for exploration is necessary.

    %In path planning for Mars rovers, engineers meplan scientific routs in an involved, time-intensive process. This process include considerations such as the 5-20 minute communication delay \cite{ormston_2012} between Earth and Mars to ensure updates to the plan are regularly provided. 
    %While during the communications delay,  
    % To reduce the amount of direction communication between Earth and rovers, 
    %NASA scientists have implemented autonomous navigation software that allows rovers to identify simple hazards over short expanses of terrain and circumvent these obstacles. 
    %However, the regions and features of interest a rover visits are still planned by scientists on Earth using images received from rovers of nearby terrain. To enable rovers to explore larger regions of minimally-charted terrain, local, image-guided trajectory planning for exploration is necessary. 
    %we propose a trajectory-planning method that utilizes image recognition to identify targets of search.

    In exploration over unknown environments, regions of interest are identified over time given new information rather than known \emph{a-priori}. As such, it is important that the rover not only prioritize obtaining new information, but also explore the unknown.  
    %balances the discovery of through  information maximization and coverage. 
    This way, the rover may further collect valuable information in regions that have already been encountered while also discovering new regions that may be overlooked. Additionally, the rover should be able to further locally adapt the plan according to the discovery of new information. 
    % Furthermore, rather than planning only one trajectory that is followed for the duration of the exploration mission, the path planning method ought to be able to calculate a new trajectory each time the information known about the space changes. 
    % This way, the newest trajectory will always distribute time with consideration to where targets have already been found. 
    Recent methods in ergodic exploration, often referred to as ergodic trajectory optimization (ETO), have shown promise in enabling robots to generate exploratory patterns that balance time spent exploring new areas and visiting known informative regions~\cite{bourgault2002information, mavrommati2017real, beinhofer2013effective,  feder1999adaptive, lerch2023safety, Dong-RSS-23}.
    In ETO, trajectories are generated such that the amount of time a robot spends in different regions of a workspace is proportional to a probability density map describing the distribution of information in that space~\cite{7350162}. 
    % Given these requirements, we found the most appropriate path-planning method to be ergodic trajectory optimization (ETO). In The density map can be updated with information collected from the sensors of an agent. 
    % So, in exploration spaces about which little-to-no initial information is known, such as an uncharted region of Mars, an initial trajectory can be planned and as the agent follows that trajectory, it can collect information used to update its belief map and then recalculate an improved trajectory. 
    % Some prior works have calculated expected measurement utility and then updated this belief map through the use of Fischer information and mutual information  However, we begin with a more simplistic, heuristics-based method: the probability density map is represented as a normalized matrix with the same dimensions as the exploration space, and then, each time a target is identified by the rover as it follows its current trajectory, the value of this matrix at the index corresponding to the location of identification is multiplied by a scalar and the matrix is then re-normalized. A new trajectory is then calculated for the updated map.
    
    \begin{figure}
        \centering
        \includegraphics[width=\linewidth]{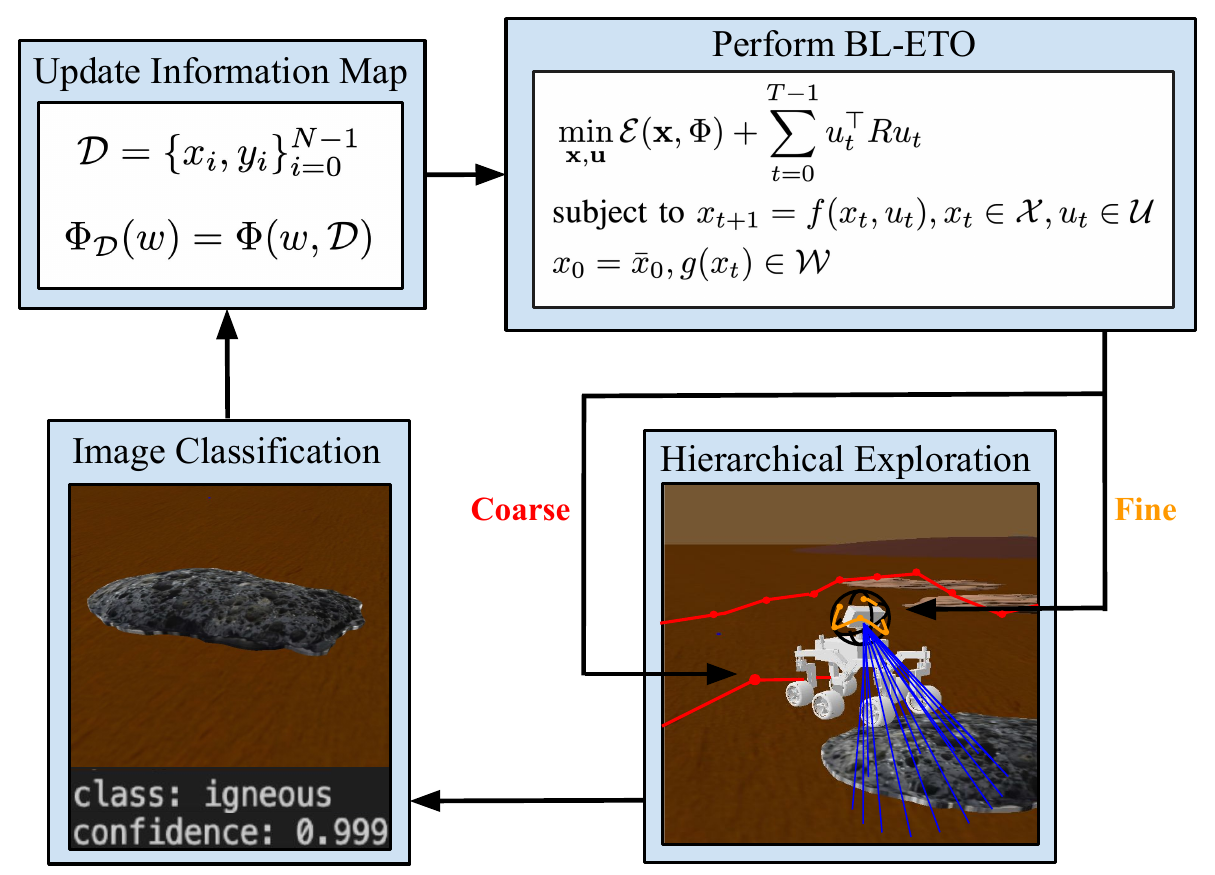}
        \caption{\textbf{Image-Based Bi-Level Ergodic Exploration.} Shown is a flow graph of our bi-level exploration process. The body trajectory and camera head trajectory are optimized separately in problems referred to as Bi-Level Ergodic Trajectory Optimization (BL-ETO). The rover main body and head follow the search trajectories with low-level controllers and takes images of the nearby terrain. The terrain is classified by a trained image identification program. We leverage coarse and fine exploration to explore and obtain useful measurements that can be overlooked.}
        \label{fig:fig 1}
    \end{figure}

    \begin{figure*}[ht]
        \includegraphics[width=\linewidth]{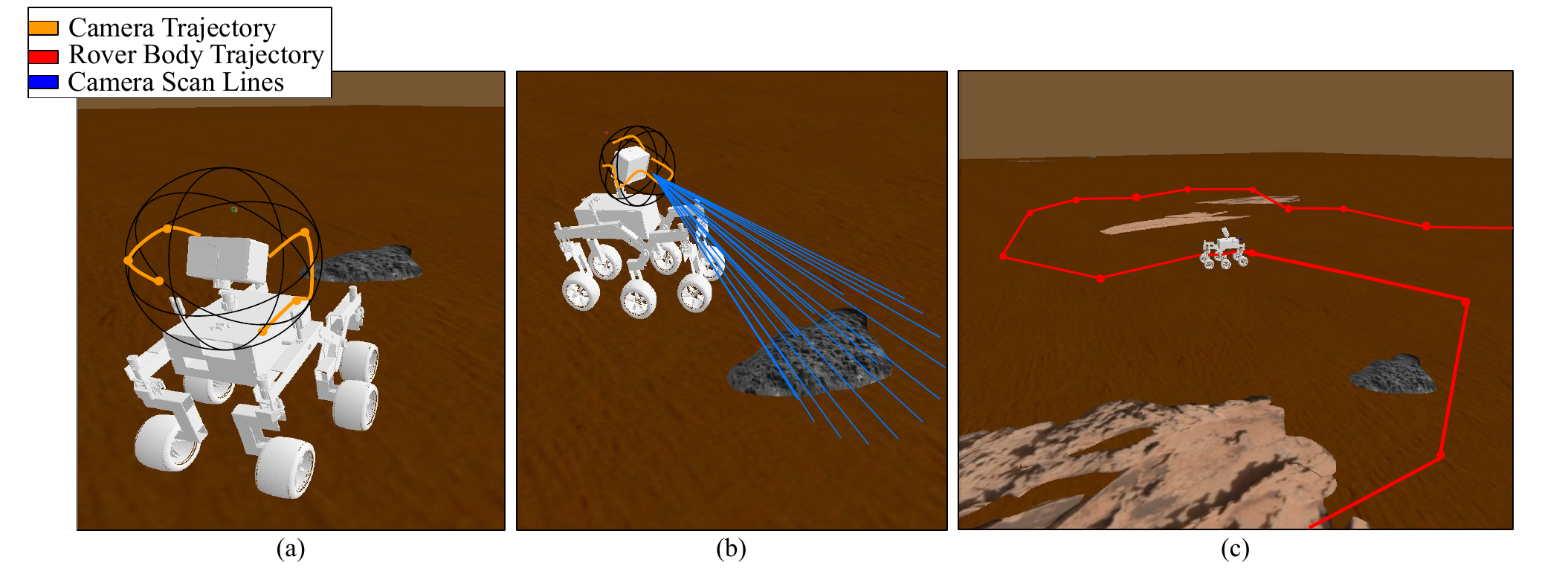}
        \caption{\textbf{Mars rover exploring for geological data with coarse and fine ergodic exploration.} For the application of a Mars rover, we show dual trajectory optimization of the rover body and the rover's visual sensor. In (a), a series of trajectory steps, mapped along the inside of a sphere, are determined for the visual sensor to follow. In (b), the visual sensor of the rover travels one step and images a region that contains a target. In (c), the trajectory followed by the body of the rover for a limited search region is plotted over the Mars terrain. This process is described quantitatively in Algorithm 1.}
        \label{fig2}
    \end{figure*}
    The integration of image-based exploration in robotic exploration is made particularly challenging by the issue of visibility. If an object of interest appears towards the edge of an image, or only a small portion of the object is visible, it is likely that the robot will fail to identify the target. Intelligent repositioning of the camera pose therefore becomes a necessary capability for vision-guided mobile robots. It is thus important both camera orientation and body position are considered in ETO.
    %, as shown in Fig. 1. When the camera follows its own optimal trajectory, it can image the nearby terrain multiple times at angles that are more likely to yield complete views of a target, enabling the image recognition software to correctly identify the target. 
    However, with increased dimensionality, ETO-based methods tend to scale poorly in terms of computational speed \cite{mavrommati2017real}. While some methods do exist to combat the computational overhead, they require pre-processing of a-priori information maps \cite{shetty2021ergodic}, which are not always available until the robot begins exploration, especially in image-guided exploration.  
    % What's more, if we attempt to optimize the trajectory of the robot position and the visual sensor orientation simultaneously, we encounter another issue: the visual output of the robot's camera changes constantly as the robot's base moves, meaning the information map of the visual sensor is changing throughout the optimization of the camera's trajectory, making the problem intractable. Between these two issues, it becomes clear that we cannot simply apply ETO to both the robot body and the visual sensor simultaneously. %utational work required to perform ETO scales exponentially with the degrees of freedom of the states \cite{9312988}. Adding two new states to the trajectory optimization- camera pitch and yaw- therefore increases the optimization run time significantly. Further, if the rover's camera and body trajectories are optimized and run together, the view output of the camera will change with each trajectory step. This means the information map of the visual sensor will constantly be changing, making the problem intractable. Since optimization over these 4 states is therefore infeasible, we propose the novel approach of decomposing the ergodic trajectory optimization into two separate optimizations that run linearly: a rover body optimization and a camera optimization. This approach keeps computational work within a feasible range while also still addressing the vision problem.
    In this work, we decompose the ergodic exploration method in a bi-level optimization, moving the body then the camera position. Splitting the optimization into two components, fine and coarse exploration, compensates for computational overhead. Each component is directly linked to separate information maps derived from the learned image-based system. 
    As such the contributions of this paper are as follows: 
    \begin{enumerate}
        \item The integration of a learned, image classifier as a guide for ergodic exploration; 
        \item A bi-level decomposition of the ergodic exploration methods with image-guided information maps; and
        \item Illustration of our approach on a simulated Mars rover geological survey environment using geological data sets.
    \end{enumerate}

% The primary contribution of this work is a decomposed, linearly-executed method of trajectory optimization used to address the issue of vision in target localization. Manipulating robot movement so that the visual sensor achieves a specific pose, thus ensuring the visual data captures a desired target, is a complex process that often requires additional sensor information or initial information on the search space and targets. Optimizing over visual sensor trajectory instead provides a robot with multiple chances to capture desired target data without requiring information beyond what is provided by the visual sensor.

\section{RELATED WORK}

\noindent
\textbf{Visual-Guided Exploration: }
    A number of different works have used visual sensor information to solve mapping, localization, and navigation problems. Prior mapping approaches have characterized local \cite{1302485, 1544951} or global \cite{7989235, iyer2018geometric} visual data through image classification technology to create maps of an environment. Localization, in which the position and orientation of a robot or a target are repeatedly estimated \cite{CEBOLLADA2021114195}, has widely been handled using visual sensor input \cite{lee2009mobile, sabattini2012experimental, chae2005combination, dellaert1999using}. Navigation tasks using visual sensor information have ranged from obstacle avoidance and safe navigation \cite{5109666, murray2000using} to information maximization \cite{davison1999mobile, martinez2009bayesian}. 
    
    Generally, in vision-guided exploration, visual data provides pose information that dictates the next position to which a mobile robot will travel. In frontier-based methods \cite{4058690}, the boundaries of the constructed map determine where the mobile robot should next visit, prioritizing capturing images of previously uncharted terrain. In these methods, visibility is a vital limitation to consider. Some common ways of addressing visibility constraints in mapping and path planning include using range sensors to determine potential occlusions \cite{saulnier2020information}, analyzing the geometry of a known map of the exploration space \cite{5109666}, or avoiding regions where sensor obscuration is predicted \cite{julian2014mutual}. Our approach handles the issue of visibility without the use of additional depth sensors or prior knowledge of landmarks within the exploration space, relying only on an RGB camera. In \cite{7139865}, the authors precompute the visibility of clusters of frontier voxels from different destinations, thereby avoiding the computational work of direct visibility checks. In our work, we bypass the large computational work of checking visibility directly along with the requirement to precompute visibility by planning camera paths relative to a map which encodes information about the likelihood of successfully observing a target from different orientations. We further extend past vision-guided exploration approaches by also revisiting known regions of high importance (e.g., landmarks) that are automatically encoded based on our learned image model. Furthermore, through the use of ergodic trajectory optimization, our approach ensures full coverage of the space without the need for explicit consideration of frontiers in path planning. To obtain more information from landmarks in the environment, a visual simultaneous localization and mapping (VSLAM) algorithm \cite{karlsson2005vslam} could be implemented into our exploration framework; this approach is left to future work.
    
    %The application of deep learning frameworks is also useful in extracting landmark information to determine if a location has been previously visited, a process known as closing the loop. However, even in VSLAM algorithms which utilize deep learning, loop closure detection may fail if landmarks are not re-observed every time step or a landmark is wrongly associated with one that was previously seen, which can cause the robot to misunderstand its position in space. The ergodic search method presented in this paper is advantageous in that if the robot fails to identify a target, it does not affect the robot's understanding of its location nor ability to complete its current trajectory step, which is pre-planned based on the current information map.

    %solves the planning problem under a single optimization problem that also guarantees complete coverage of a space with enough time~\cite{++}

\vspace{1mm}
\noindent
\textbf{Ergodic Exploration: }
    As described previously, ergodic exploration is a recent method for distributing coverage of a region proportional to the expected information within that region. In many prior works on ergodic trajectory optimization, data collected during exploration is used to update a non-stationary information map \cite{coffin2022multi, mavrommati2017real}.
    % if the information density map is expected to change regularly, model predictive control (MPC) is commonly applied . In MPC, at every time step, a new control action that seeks to improve ergodicity optimally is calculated \cite{mavrommati2017real}. MPC is therefore highly computationally expensive. In applications where micro-adjustments must be made to the state of agents in response to every sensor reading, MPC is necessary for the function of the system and focus is then shifted to decreasing the computational cost to a feasible range, as in []. 
    Prior work has often been limited to a minimal number of robot states to mitigate computational overhead. However, our work mitiages computational overhead not by limiting robot states, but rather by decomposing the optimization problem. Other approaches have developed sample-based techniques and formulations that reduce the computational cost \cite{9312988}. These require assumptions on prior information that cannot be obtained immediately in real-time image-based exploration, such as the distribution of all landmarks in the space. In addition, sample-based approximations do not guarantee complete coverage and are limited numerically to the scale of the exploration space, whereas the canonical ergodic exploration formulation is considered multi-scale, i.e., the exploration space can be arbitrarily large or small \cite{mathew2011metrics}. This work integrates a learned image-classifier and leverages an occupancy grid to project information to the robot's future poses, while maintaining the multi-scale capabilities of the ergodic formulation to the extent of the memory requirements of the occupancy grid.
    
\section{METHODS}
In this section, we describe the ergodic trajectory optimization method and extend the method to image-based information maps with bi-level optimization.

 \subsection{Ergodic Trajectory Optimization}
        Let us define the state the robot at some discrete time $t$ as $x_t \in \mathcal{X} \subset \mathbb{R}^n$ and the control input to the robot as $u_t \in \mathcal{U} \subset \mathbb{R}^m$. 
        In addition, let us define the robot's workspace as $\mathcal{W} \in [0,L_0]\times \ldots [0, L_{v-1}]$ where $L_i$ are the bounds of the space and $v\leq n$ is the dimensionality. We also define a map $g: \mathcal{X} \to \mathcal{W}$ which is continuous and differentiable and maps a robot's state $x_t$ to a point in the exploration space (e.g., Euclidean space); that is, $g(x) = w$ and $w \in \mathcal{W}$.
        A trajectory of the robot $x_{0:T-1}$ for some time horizon $T$ is given by the equation
        \begin{equation}
            x_{t+1} = f(x_t, u_t)
        \end{equation}
        where $f: \mathcal{X}\times \mathcal{U} \to \mathcal{X}$ are the motion constraints of the robot and $x_{0:T-1}$ is given by recursive application of $f(x,u)$ from some initial condition $x_0$.
        
        Formally, a trajectory is said to be ergodic if its \emph{time-averaged statistics}, that is, the amount of time a robot spends over a workspace $\mathcal{W}$, is proportional to some measure of information $\Phi : \mathcal{W} \to \mathbb{R}$ over the workspace. \footnote{The measure $\Phi$ can encode any information over the space $\mathcal{W}$, e.g., uncertainty, and it follows that $\int_\mathcal{W} \Phi(w) dw = 1$ and $\Phi(w) \neq 0 \forall w \in \mathcal{W}$.} 
        For a deterministic trajectory $x_{0:T-1}$, we define ergodicity as 
        \begin{equation} \label{eq:ergodicity}
            \lim_{T\to \infty} \frac{1}{T}\sum_{t=0}^{T-1} h(g(x_t)) = \int_\mathcal{W} \Phi(w)h(w) dw
        \end{equation}
        for all Lebesque integrable functions $h \in \mathcal{L}^1$ \cite{de2016ergodic}.

        We optimize trajectories $\mathbf{x} = x_{0:T-1}$ and control signals $\mathbf{u} = u_{0:T-1}$ to minimize an error metric we define, called the \emph{ergodic metric}:
        \begin{align}
            &\mathcal{E}(\mathbf{x}, \Phi) = \sum_{k\in \mathbb{N}^v} \Lambda_k \left( c_k(x_{0:T-1}) - \Phi_k \right)^2 \\
            &= \sum_{k\in \mathbb{N}^v} \Lambda_k \left( \frac{1}{T}\sum_{t=0}^{T-1}F_k(g(x_{t})) - \int_{\mathcal{W}} \Phi(w) F_k(w)dw \right)^2 \nonumber,
        \end{align}
        where $F_k(w) = \prod_{i=0}^{v-1} \cos(w_i k_i \pi/ L_i)/h_k$ is the cosine Fourier transform for the $k^\text{th}$ mode, $\Lambda_k$ is a weight described by \cite{mathew2011metrics}, and $h_k$ is a normalization factor~\cite{7350162}. Fourier coefficients are used so that the metric is differentiable with respect to the trajectory $\mathbf{x}$ and thus can be minimized. Some common alternative methods to compare the spatial distance between distributions, like Kullback-Leibler divergence \cite{kullback1951information} and the Earth-mover's distance \cite{rubner2000earth}, are not necessarily differentiable with respect to trajectory. Thus, the ergodic metric is preferred for this application. We construct an augmented cost function composed of the ergodic metric $\mathcal{E}$ and the control input to the system $\mu_{t}$:
        \begin{align} \label{eq:erg_opt}
            & \min_{\mathbf{x}, \mathbf{u}}\mathcal{J}(x, t)= \mathcal{E}(\mathbf{x}, \Phi) + \sum_{t=0}^{T-1} u_t^\top R u_t \\ 
            &\text{subject to } x_{t+1} = f(x_t, u_t),         x_t \in \mathcal{X}, u_t \in \mathcal{U} \nonumber \\ 
            & x_0 = \bar{x}_0, g(x_t) \in \mathcal{W}  \nonumber
        \end{align}
        where $\bar{x}_0$ is an initial condition and $R$ is a weight controlling the relative priority of minimizing distance from ergodicity. 
        We solve Eq.~\eqref{eq:erg_opt} as a transcription-based nonlinear program using an interior point constraint solver~\cite{boyd2004convex,lukvsan2004interior}.

\subsection{Image-Guided Information}
    Integrating image-data into the ergodic formulation requires that we leverage the existing structure of the algorithm. Specifically, we note that $\Phi$, the information measure, can be defined arbitrarily so long as it is defined over the exploration space $\mathcal{W}$ and holds the properties of being positive definite $\Phi(w)>0 \forall w$ and $\int_\mathcal{W} \Phi(w) dw = 1$.
    To accomplish this, let us define the $\Phi$ as a function of past robot states $x_i$ and collected image labels $y_i$. Let $y = \gamma(w)$ be a model of a learned image classifier such that $y \in \mathcal{Y}$ is a one-hot encoded label and $\gamma : \mathcal{W} \to \mathcal{Y}$ is a mapping function that generates an image label given the current pose (i.e. location and orientation) of the robot in the exploration space.\footnote{This function is used for notation and can only be evaluated experimentally with data and defined by the environment.  }
    
    Given a collected data set of $N$ points, $\mathcal{D} = \{ x_i, y_i \}_{i=0}^{N-1}$, we have $\Phi_\mathcal{D}(w) = \Phi(w, \mathcal{D})$, which we define using the map $\gamma^{-1}(y) = w$. 
    The inverse of $\Phi$ is used to map observed objects in the image space, which were classified into poses $w$, through the use of a camera pin-hole model into a point in the robot's exploration space. Therefore, from collected data, we use an occupancy map \cite{elfes1989using} in the exploration space and define $\Phi$ as a function of detected objects. 
    
    The forward function $\gamma$ is obtained by training an image classifier on a data-set $\mathcal{D}$ and is dependant on the environment the robot is exploring (which the robot does not know \emph{a-priori}). The inverse model is obtained from the camera parameters used in simulation. 

\subsection{Bi-Level Ergodic Trajectory Optimization} 
    With the means to generate ergodic trajectories given collected image-data through $\Phi$, we now sub-divide the trajectory to improve computation. 
    First, we establish a notion of \emph{coarse} and \emph{fine} exploration. 
    For a mobile robot that is tasked to explore a large region, we say the robot is equipped with a camera that can freely rotate, but is attached to the body (i.e., the camera can not displace itself further than the robot's body pose). 
    This decomposition has the advantage of allowing the ergodic planner to generate trajectories that span a larger area while ensuring locally the robot is finely observing with several camera images. 

    Next, let the body pose of the robot be defined over a subspace $\mathcal{W}_c \subset \mathcal{W}$ and the configuration of the camera over the subspace $\mathcal{W}_f \subset \mathcal{W}$. 
    Likewise, let $g_c: \mathcal{X} \to \mathcal{W}_c$ and $g_f : \mathcal{X} \to \mathcal{W}_f$ be functions that pull the relevant robot states $x$ to the coarse and fine exploration spaces.
    Note that while the two subspaces are separate, the elements in $\mathcal{W}_f$ change as the robot moves. To account for this, we choose to ``freeze'' the information that the camera sees so that planning is tractable.
    %Not doing so will present itself to a highly coupled and intractable problem due to predictive planning. 
    
    The optimization described in \eqref{eq:erg_opt} is then subdivided into two components: 
    \begin{align}
            &\min_{\mathbf{x}_c,\mathbf{u}_c} \mathcal{E}(\mathbf{x}_c, \Phi_c) \text{ and } \min_{\mathbf{x}_f,\mathbf{u}_f} \mathcal{E}(\mathbf{x}_f, \Phi_f) \nonumber
    \end{align}
    where the subscript defines the fine and coarse dimensions, the dynamics are assumed to be decoupled into $f_c(x_c, u_c)$ and $f_f(x_f, u_f)$,\footnote{For a kinematic system, the chain is broken into sub-components that are assumed independent of one another. } and the cost on control effort is omitted for clarity. The fine optimization problem is a function of the coarse optimization problem, meaning fine trajectory optimization occurs in full during each step of the coarse trajectory. The decomposition of the information map are given by the inverse map $\gamma^{-1}$. An illustration of our combined approach is shown in Fig.~\ref{fig:fig 1} and described in Algorithm~\ref{alg:1}.
\vspace{3mm}
    
\begin{algorithm}
 \caption{Bi-Level, Image-Based Ergodic Exploration}
 \begin{algorithmic}[1]
  \State \textbf{init:} $\Phi_c, \Phi_f$, terminal discrete time $t_f$
  \While{$t<t_f$}
    \State $p \gets$ getRobotBasePose()
    \State $\mathbf{x}_c, \mathbf{u}_c =$ ergodicCoarsePlanner($p$, $\Phi_c$)
    \State Execute coarse plan
    \For{ $x_{c,i},u_{c,i} \in \mathbf{x}_c, \mathbf{u}_c$}
        \State $\theta, \psi \gets$ getRobotCameraPose()
        \State $\mathbf{x}_f, \mathbf{u}_f \gets$ ergodicFinePlanner($\theta,\psi$, $\Phi_f$)
        \For{$x_{f,j},u_{f,j} \in \mathbf{x}_f, \mathbf{u}_f$}
            \State take image and get label $y$
            \State update($\Phi_c$, $\Phi_f$)
            \State step robot camera pose
        \EndFor
        \State step robot body pose
        \State $t \gets t + 1$
    \EndFor
  \EndWhile
  \end{algorithmic} \label{alg:1}
\end{algorithm}

\section{SEARCH FOR ROCKS WITH A MARS ROVER}
\subsection{Simulation Environment \& Geological Data-Set}
    We developed a simulated Martian environment with a functional Mars rover using PyBullet \cite{coumans2021}. The rover is made to replicate the NASA JPL open source rover and is designed by \cite{rover}. 
    % We used PyBullet as our simulation environment, and for the rover, we used a URDF file modeled after the Open Source Rover from NASA JPL, made by \cite{++}. The motion of the six-wheeled rover is controlled by corner wheel axle angle and wheel velocity; the control of the rover is similar to that of a car. However, rather than using the dynamics of a kinematic car in trajectory optimization, we instead treat the rover as a point mass for calculating the trajectory as this more simplistic dynamical system is less computationally intensive. As previously described, we created a program that can rotate and move the rover from its current position $(x_i, y_i)$ to a target coordinate $(x,y)$. With this program, our program maintains realistic rover movement while also allowing for the use point mass dynamics in ETO. 
    We constructed a data-set of sedimentary and igneous rocks; these specific rock types were chosen as they both contain geological information useful in the search for conditions on Mars that were once habitable to life \cite{MANGOLD20171}.
    The data-set was obtained from NASA using a repository of images from Martian rovers and satellites called the PDS Imaging Atlas. We created 21 models using image textures for sedimentary and igneous rocks to add to the terrain environment, which was done at random using a uniform distribution. A simulated $100m^2$ search area is used for the rover to explore. The environment includes non-uniform terrain generated randomly with a bias towards predetermined hills. 
    
    The convolutional neural network (CNN) used for target classification was trained using Keras, a Tensorflow interflace \cite{tensorflow2015-whitepaper}. The CNN was a sequential model with three classes: 'igneous', 'sedimentary', and 'background'. The background class referred to any image without visible rocks, and was trained on real and simulated images of diverse Mars terrain containing plains, hills, and arches. The two rock classes were also trained on 150 images each of both real and simulated Mars rocks. Many of the training images contained rocks which were partially out-of-frame to ensure the robot could detect rocks on the edges of its vision. During training, our model had an accuracy of $99.37\%$ for the images on which it was trained and an accuracy of $97.30\%$ on a separate set of validation images. The average inference time for the CNN was $0.139s \pm 0.021$, enabling classification to occur in real-time.

\subsection{Coarse Exploration} 
    We broadly outline the steps of decomposed coarse and fine search with regards to our Mars rover application. First, an exploratory trajectory is calculated for the rover body to follow. The rover travels one step of that trajectory, then stops and calculates a trajectory for the camera to follow. The camera takes an image after each step in the camera trajectory, and if a rock is identified following any one of these steps, the information map of the camera is updated and a new camera trajectory is calculated. After the camera has finished following its trajectory, if one or more rocks has been identified, the information map of the rover body is also updated according to this identification, and a new trajectory is calculated for the rover body. The rover body then moves along the next step of its trajectory, and this process is repeated. To understand this process quantitatively, we first create an information map for coarse exploration.

    The coarse (i.e., rover body) information map $\Phi_c$ is represented as a discretized occupancy grid matrix over the exploration space. 
    To create our initial information map $\Phi_{{c}_{i}}$, we first normalize a uniform $100 \times 100$ grid matrix. 
    We initialize $\Phi_{{c}_{i}}$ as a uniform distribution that is augmented with each successive measurement of a rock with a probability measure $p(y \mid \Phi, x)$, where $\Phi$ is the respective information map and is normalized accordingly. We then scale the value of the information map in two epicenters of regions estimated to have above-average concentrations of targets.
    % We then identified two regions of the exploration space with above-average concentrations of targets and scaled the value of $\Phi_{{c}_{i}}$ in these regions by a factor of 50. 
    This adjustment to the otherwise uniform map encourages the rover to visit general areas of geological interest without any specific information about the location of individual rocks.
    % thus testing if the ergodic algorithm would be able to find small targets using nonspecific preliminary information. 
    Next, a trajectory is generated over a planar terrain using a kinematic unicycle motion model for planning. 
    A maximum allowable change in position is added as a constraint and planning is done over a time horizon of $T=48$ discrete time steps, which we found empirically to work well.
    % plane for the rover body to follow. We apply inequality constraint that each step of the trajectory is equal in magnitude. With a time horizon of $T=48$, we recalculate the states and controls for 300 iterations to generate a complete trajectory. 
    As the dynamics of the rover, which are similar to that of a kinematic car, must be considered in driving the rover to a target position, a low-level controller calculates necessary wheel orientations and velocities given a target coordinate input and current position. After calculating the coarse trajectory, we then begin the fine exploration process with the mounted camera.

\begin{figure}[b!]
    \centering
    \captionsetup{width=.45\linewidth}
    \subfloat[The surface on which the trajectory of the camera will be generated is a sphere with a segment removed in the area occupied by the body of the rover.]{\includegraphics[width=.5\columnwidth]{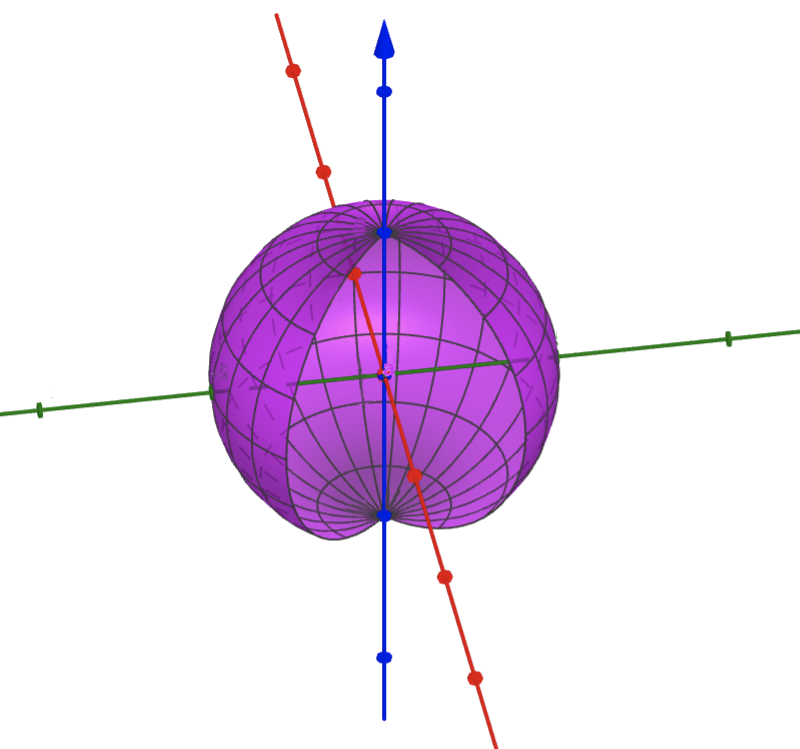}}
    \subfloat[From above, the yaw $\theta$ of the camera is $0^\circ$ when looking straight outwards from the rover body, increases in the clockwise direction, and decreases in the counterclockwise direction.]{\includegraphics[width=.5\columnwidth]{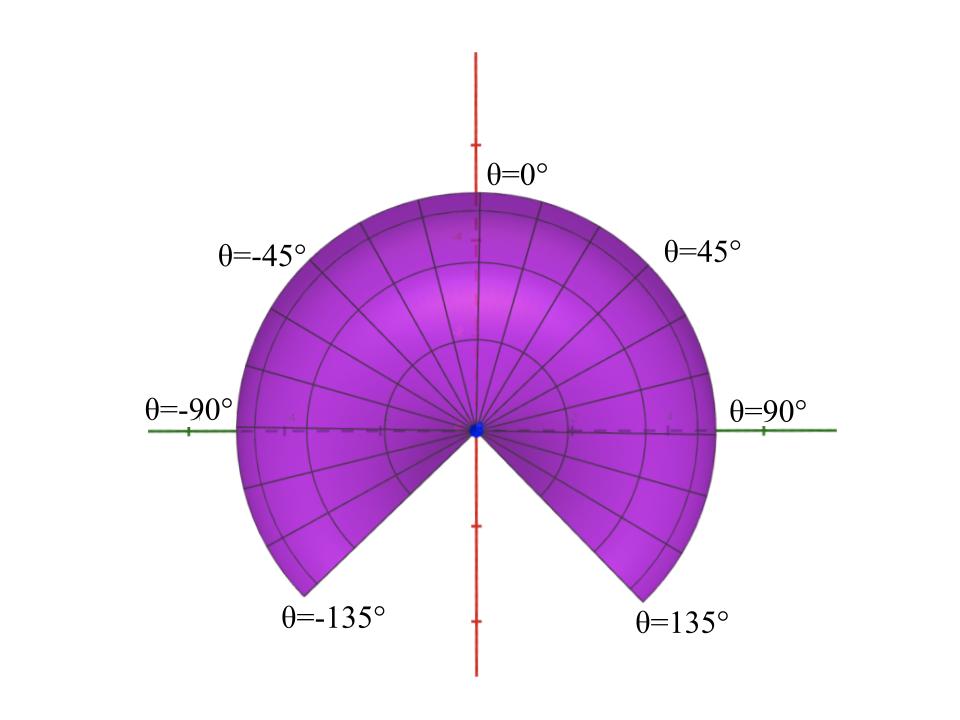}}
    \captionsetup{width=\linewidth}
    \caption{The restricted view output of the camera, pictured from the side in (a) and from above in (b).} 
    \label{fig:spherical_space}
\end{figure}

\subsection{Fine Exploration} 

    We consider an $\mathcal{S}^2$ spherical space on which the camera trajectory will be generated. 
    % The camera has a view output of a sphere and is directed to focus on the point .005m away from where it is facing, meaning the target position of the camera, $p$, has a fixed radius $r=.005$. 
    The two variables that can be altered to change the orientation of the camera are pitch $\psi$ and yaw $\theta$. The default orientation of the camera looks straight outward from the rover at $(\theta, \psi)=(0, 0)$. 
    The occlusion from the rover body is removed from the spherical exploration space which renders the space bounded between $(-135^\circ, 135^\circ)$.
    An illustration of the spherical exploration space is shown in Fig.~\ref{fig:spherical_space}.

    We then consider how to define the fine (i.e., camera) information map, which we will call $\Phi_{f}$. Given that the camera orientation optimization only considers two variables, $\theta$ and $\psi$, we define $\Phi_f$ to be a discretized two-dimensional matrix representing a projected occupancy grid in $\theta$ and $\psi$.
    % each row represents $\theta_i$ and each column represents $\psi_j$. Both $\theta$ and $\psi$ will change in increments of $1^\circ$, so $\Phi_f$ is of size $270 \times 180$. 
    To initialize and update $\Phi_f$, we note that at $\psi =0^\circ$, the camera points at the horizon. 
    From this position, we project the information map from the field of view onto the global position on the surface.
    Since rocks sit on the plane, the camera is only likely to identify rocks when looking downwards ($\psi<0$) from its default position. Following each image classification, the occupancy map is updated based on whether a rock was identified. The resulting map is then normalized and used to generate an ergodic trajectory through the fine trajectory optimization in Eq. \eqref{eq:erg_opt}.
    % Looking further downwards $\psi<-30$ increases the odds that a rock sitting on the plane will be in full view. However, when $\psi < -50$, the camera points too far downwards and is able to miss rocks that are only a couple of meters away. So, for a uniform, normalized $\Phi_{f_{i}}$, we multiply the indices in the range where $-30<\psi <0$ by 500 and the indices in the range where $-50<\psi<-30$ by 2000, and finally re-normalize $\Phi_{f_{i}}$. Equipped with an initial information map, we now may generate a trajectory for the camera to follow.

    We set the camera time horizon $T=5$ discrete time steps, meaning 5 images ranging across the surrounding terrain will be captured and classified. Movement on the camera is restricted through inequality constraints and modeled using a single-integrator motion model. The position of the camera is updated using a low-level controller that tracks the exploration plan. An image is captured and classified at each individual step during trajectory execution. 
    % Additionally, we again define inequality constraints that force each trajectory step to be equal in magnitude, ensuring that the full trajectory will cover a large window of view. We first generate a trajectory, calculated over 300 iterations, over the initial camera information map $\Phi_{f_{i}}$. At each step $s$ of the trajectory, the camera target position $P_s$ is updated to 
    % \begin{equation}
    % P_s=(.005, \theta_i + \theta_s,\psi_i + \psi_s),
    % \end{equation}
    % where $\theta_i$ and $\psi_i$ are the initial yaw and pitch of the camera. At any step $s$, if a rock is identified, the value of the information map at $\Phi_{{f}_{\theta \psi}}$ is scaled by 500 and a new trajectory with $T-s$ steps is calculated. 
    If a rock is identified, both the fine information map and the coarse information map are updated accordingly.
    % After the camera has followed its trajectory steps, the information map corresponding to the rover body, $\Phi_c$, is updated. If any rocks were discovered in the images taken, $\Phi_c$ at the location of identification, $(x,y)$, is scaled by 10. 
    To prevent repeated spiking of the information map in one location in the case that one rock is encountered multiple times, scaling of the information map upon identification is clipped if a prior identification was made at that same location.

\begin{figure*}[t!]
    \centering
    \includegraphics[width=\linewidth]{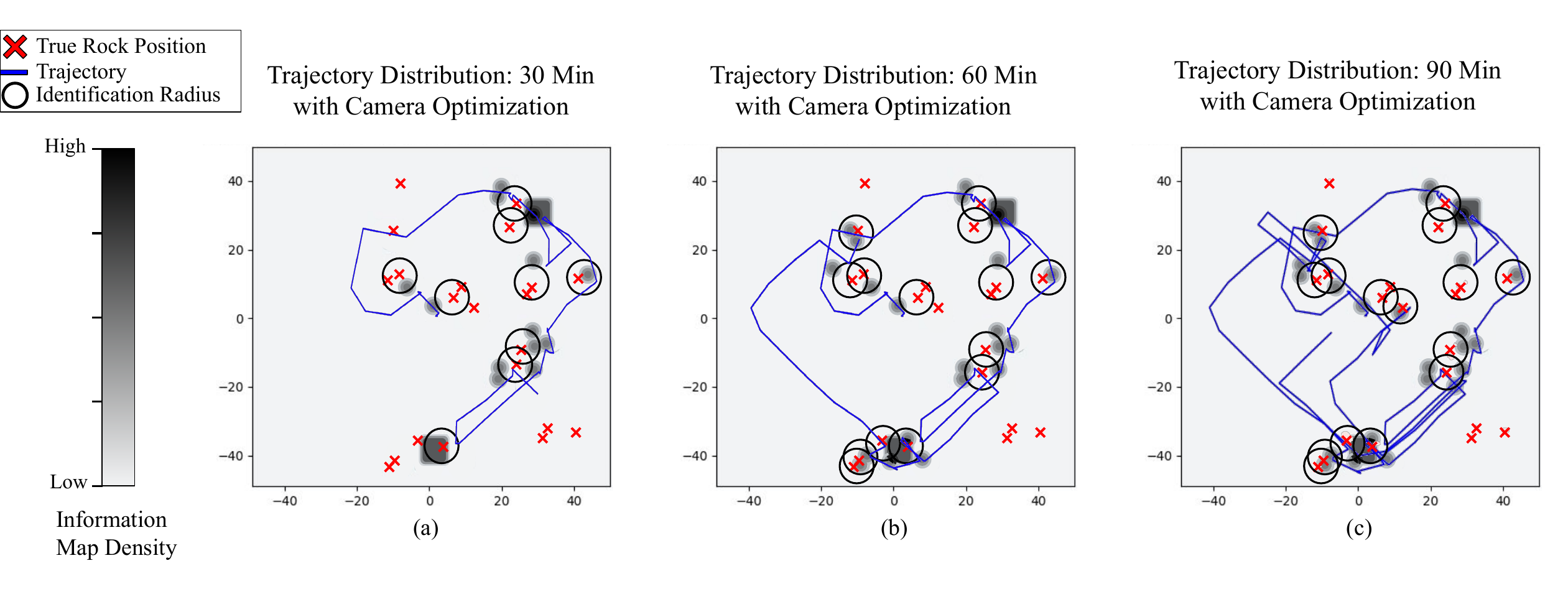}
    \caption{The trajectory is plotted over 30, 60, and 90 minute time intervals from a top-down view of the exploration space. The information density map is plotted in gray scale. The small circular regions where the information density increases rapidly indicate points where the rover identified targets, while the larger rectangular regions mark the areas at which the initial information map was increased. The red Xs mark the true locations of rocks in the exploration space. Experimentally, we determined that the rover is able to identify rocks from at most about 5 meters away. To calculate how many rocks were discovered, we overlay black circles of radius 5 meters atop the true locations of the rocks; if a spike in the information map falls within one of these circles, it indicates that the corresponding rock was identified.}
    \label{fig:rover_explr}
\end{figure*}

\section{RESULTS}

The results are used to demonstrate the effectiveness of our approach for discovering rocks in the environment through our BL-ETO method. We demonstrate that our method can be used to discover a higher portion of rocks in the environment through optimization of the camera trajectory. Furthermore, we present comparisons with other naive approaches to show that our approach can plan effective exploratory trajectories with non-stationary information maps that are updated in run time. 

\subsection{Rocks Identified with Optimized Camera Orientation}
    We first performed a simulation in which both the rover body and camera trajectories are optimized. The simulation is run over the course of 90 minutes, with the trajectory of the rover plotted in Fig. \ref{fig:rover_explr} at benchmarks of 30, 60, and 90 minutes. We observe that the trajectories are ergodic with respect to the coarse information map $\Phi_c$ as the space is well-covered yet more time is spent near the two regions of initial high initial information. 
    
    With each new discovery of a rock, a data-point is appended and updates the information map as shown in Fig.~\ref{fig:rover_explr}. 
    The dark regions of the information map correspond with the true location of rocks (shown as a red cross).  As the rover identifies rocks in particular regions, depicted by the dark circles, the information map gets updated during run time. To determine the number of rocks discovered by the end of the simulation, we compare the locations of identification with the true rock locations. Using this method, the rover found, on average, $58\%$ of the rocks in the space over a 90 minute span.

\subsection{Comparisons to Naive Baseline Methods}

\begin{table}[b!]
\begin{tabular}{l|l|l|l}
\hline
Method                 & \begin{tabular}[c]{@{}l@{}}Avg Rocks\\ Found\end{tabular} & \begin{tabular}[c]{@{}l@{}}Std Deviation\\ Rocks Found\end{tabular} & \begin{tabular}[c]{@{}l@{}}Length of\\ Path\end{tabular} \\ \hline
ETO Fixed Camera       & 30.16\%                                                   & 8.09\%                                                              & 1576.2 m                                                 \\
ETO Random Camera Path & 23.81\%                                                   & 5.83\%                                                              & 1598.2 m                                                 \\
BL-ETO Optimized Camera & 58.10\%                                                   & 9.23\%                                                              & 755.7 m                                                 
\end{tabular}
\caption{Comparison results of H-ETO with ETO using a fixed and random camera path. H-ETO is capable of identifying more rocks due to the fine exploration while minimizing the path length traveled during exploration. }\label{tab:comp}
\end{table}

    We first compared our method to regular ergodic trajectory optimization with the visual sensor fixed. This is the baseline method of navigation approaches that treat a robot's camera and body as one unit, controlling camera head movement through rotation of the body. Additionally, we compared our method to an ergodic trajectory optimization in which the visual sensor moved according to a random trajectory. In this approach, to make the comparison as fair as possible, we limit the set of potential coordinates chosen to exclude the occlusion from the rover body and ensured that an equal number of images were taken during the camera pan as in the optimal method.
    % As the precedent for search and target localization with visual sensors is for the sensor to be fixed with respect to the body of the robot, we also ran a simulation in which the camera is fixed.
    Since the fixed camera only observed the forward position, it effectively had a single camera image to use for identifying the rocks. 
    % Additionally, we recognized that as the optimized camera trajectory method took 5 images as it followed a path of 5 trajectory steps, the fixed camera method disadvantageously only took 1 image. 
    % So, as a second baseline against which to compare, we ran a simulation in which the camera moves according to a randomly-generated trajectory with an equal number of steps as the optimized trajectory. 
    A comparison of these three methods is summarized in Table~\ref{tab:comp}. 
    % Both the randomly-generated trajectory method and the fixed method achieved better coverage than the optimal trajectory method. 
    % This can be attributed to the amount of time required for path planning and classification for each method, which is also described in table 1. 
    % It takes about 1.5s to capture and  classify 1 image, meaning that to capture and classify 5 images in the random trajectory method took about 7.5s. Only in the optimal trajectory method was time spent on path planning; together, calculating an optimal path and classifying/capturing 5 images took about 22s. 
    We compare the total path length explored by the rover in each method. Intuitively, the ETO with fixed and random camera approaches had longer path lengths as less time is required for image capture and identification when a new camera trajectory does not regularly have to be calculated. Yet while collectively, the proposed BL-ETO method required more computation, the approach was able to identify significantly more rocks in the environment compared with the other methods. In particular, over five trials, BL-ETO found $58\%$ of the total rocks in the environment where ETO with a fixed camera and ETO with a randomized camera path detected $30\%$ and $24\%$ of rocks, respectively. The addition of a guided camera exploration path that adapts to run-time image data adds a necessary amount of finer exploration that compensates for overlooked areas in the rover's path.

    % However, despite the improved coverage in the two baseline methods, the optimized trajectory method performed significantly better in localization. Compared to the 15 rocks identified with the optimal trajectory method, 8 rocks were identified with the fixed method, and only 5 with the random trajectory method. 
    % Though it is not initially intuitive that this imaging method would be less successful than when the sensor was fixed, when we consider that most of the potential camera orientations are not viable for capturing targets that sit on the terrain (i.e., the camera could be pointed towards the sky or looking straight down), it becomes clear that a randomly-generated camera trajectory is unlikely to succeed.

\section{CONCLUSIONS}
In this paper, we show that bi-level decomposed ergodic search is an effective method to deal with visibility limitations that arise in mobile robot navigation and localization. Our approach of optimizing over visual sensor trajectory showed significantly better performance in localization than two baseline methods. Moreover, we demonstrate that the computational cost of vision-guided ergodic exploration can be managed through decomposition of the trajectory optimization into separate groups. In future work, we would test this search approach in the real world to better understand how learned image-based classification changes on less understood, more heterogeneous objects.

\bibliographystyle{IEEEtran}
\bibliography{reference}

\begin{thebibliography}{10}
\providecommand{\url}[1]{#1}
\csname url@samestyle\endcsname
\providecommand{\newblock}{\relax}
\providecommand{\bibinfo}[2]{#2}
\providecommand{\BIBentrySTDinterwordspacing}{\spaceskip=0pt\relax}
\providecommand{\BIBentryALTinterwordstretchfactor}{4}
\providecommand{\BIBentryALTinterwordspacing}{\spaceskip=\fontdimen2\font plus
\BIBentryALTinterwordstretchfactor\fontdimen3\font minus
  \fontdimen4\font\relax}
\providecommand{\BIBforeignlanguage}[2]{{%
\expandafter\ifx\csname l@#1\endcsname\relax
\typeout{** WARNING: IEEEtran.bst: No hyphenation pattern has been}%
\typeout{** loaded for the language `#1'. Using the pattern for}%
\typeout{** the default language instead.}%
\else
\language=\csname l@#1\endcsname
\fi
#2}}
\providecommand{\BIBdecl}{\relax}
\BIBdecl

\bibitem{ormston_2012}
\BIBentryALTinterwordspacing
T.~Ormston, ``Time delay between mars and earth,'' Aug 2012. [Online].
  Available:
  \url{https://blogs.esa.int/mex/2012/08/05/time-delay-between-mars-and-earth/}
\BIBentrySTDinterwordspacing

\bibitem{bourgault2002information}
F.~Bourgault, A.~A. Makarenko, S.~B. Williams, B.~Grocholsky, and H.~F.
  Durrant-Whyte, ``Information based adaptive robotic exploration,'' in
  \emph{IEEE/RSJ international conference on intelligent robots and systems},
  vol.~1.\hskip 1em plus 0.5em minus 0.4em\relax IEEE, 2002, pp. 540--545.

\bibitem{mavrommati2017real}
A.~Mavrommati, E.~Tzorakoleftherakis, I.~Abraham, and T.~D. Murphey,
  ``Real-time area coverage and target localization using receding-horizon
  ergodic exploration,'' \emph{IEEE Transactions on Robotics}, vol.~34, no.~1,
  pp. 62--80, 2017.

\bibitem{beinhofer2013effective}
M.~Beinhofer, J.~M{\"u}ller, and W.~Burgard, ``Effective landmark placement for
  accurate and reliable mobile robot navigation,'' \emph{Robotics and
  Autonomous Systems}, vol.~61, no.~10, pp. 1060--1069, 2013.

\bibitem{feder1999adaptive}
H.~J.~S. Feder, J.~J. Leonard, and C.~M. Smith, ``Adaptive mobile robot
  navigation and mapping,'' \emph{The International Journal of Robotics
  Research}, vol.~18, no.~7, pp. 650--668, 1999.

\bibitem{lerch2023safety}
C.~Lerch, D.~Dong, and I.~Abraham, ``Safety-critical ergodic exploration in
  cluttered environments via control barrier functions,'' in \emph{2023 IEEE
  International Conference on Robotics and Automation (ICRA)}.\hskip 1em plus
  0.5em minus 0.4em\relax IEEE, 2023, pp. 10\,205--10\,211.

\bibitem{Dong-RSS-23}
D.~E. Dong, H.~P. Berger, and I.~Abraham, ``{Time Optimal Ergodic Search},'' in
  \emph{Proceedings of Robotics: Science and Systems}, Daegu, Republic of
  Korea, July 2023.

\bibitem{7350162}
L.~M. Miller, Y.~Silverman, M.~A. MacIver, and T.~D. Murphey, ``Ergodic
  exploration of distributed information,'' \emph{IEEE Transactions on
  Robotics}, vol.~32, no.~1, pp. 36--52, 2016.

\bibitem{shetty2021ergodic}
S.~Shetty, J.~Silv{\'e}rio, and S.~Calinon, ``Ergodic exploration using tensor
  train: Applications in insertion tasks,'' \emph{IEEE Transactions on
  Robotics}, vol.~38, no.~2, pp. 906--921, 2021.

\bibitem{1302485}
B.~Kuipers, J.~Modayil, P.~Beeson, M.~MacMahon, and F.~Savelli, ``Local
  metrical and global topological maps in the hybrid spatial semantic
  hierarchy,'' in \emph{IEEE International Conference on Robotics and
  Automation, 2004. Proceedings. ICRA '04. 2004}, vol.~5, 2004, pp. 4845--4851
  Vol.5.

\bibitem{1544951}
Z.~Zivkovic, B.~Bakker, and B.~Krose, ``Hierarchical map building using visual
  landmarks and geometric constraints,'' in \emph{2005 IEEE/RSJ International
  Conference on Intelligent Robots and Systems}, 2005, pp. 2480--2485.

\bibitem{7989235}
V.~Peretroukhin, L.~Clement, and J.~Kelly, ``Reducing drift in visual odometry
  by inferring sun direction using a bayesian convolutional neural network,''
  in \emph{2017 IEEE International Conference on Robotics and Automation
  (ICRA)}, 2017, pp. 2035--2042.

\bibitem{iyer2018geometric}
G.~Iyer, J.~Krishna~Murthy, G.~Gupta, M.~Krishna, and L.~Paull, ``Geometric
  consistency for self-supervised end-to-end visual odometry,'' in
  \emph{Proceedings of the IEEE Conference on Computer Vision and Pattern
  Recognition Workshops}, 2018, pp. 267--275.

\bibitem{CEBOLLADA2021114195}
\BIBentryALTinterwordspacing
S.~Cebollada, L.~Payá, M.~Flores, A.~Peidró, and O.~Reinoso, ``A
  state-of-the-art review on mobile robotics tasks using artificial
  intelligence and visual data,'' \emph{Expert Systems with Applications}, vol.
  167, p. 114195, 2021. [Online]. Available:
  \url{https://www.sciencedirect.com/science/article/pii/S095741742030926X}
\BIBentrySTDinterwordspacing

\bibitem{lee2009mobile}
Y.-J. Lee, B.-D. Yim, and J.-B. Song, ``Mobile robot localization based on
  effective combination of vision and range sensors,'' \emph{International
  Journal of Control, Automation and Systems}, vol.~7, no.~1, pp. 97--104,
  2009.

\bibitem{sabattini2012experimental}
L.~Sabattini, A.~Levratti, F.~Venturi, E.~Amplo, C.~Fantuzzi, and C.~Secchi,
  ``Experimental comparison of 3d vision sensors for mobile robot localization
  for industrial application: Stereo-camera and rgb-d sensor,'' in \emph{2012
  12th International Conference on Control Automation Robotics \& Vision
  (ICARCV)}.\hskip 1em plus 0.5em minus 0.4em\relax IEEE, 2012, pp. 823--828.

\bibitem{chae2005combination}
H.~Chae and K.~Han, ``Combination of rfid and vision for mobile robot
  localization,'' in \emph{2005 International Conference on Intelligent
  Sensors, Sensor Networks and Information Processing}.\hskip 1em plus 0.5em
  minus 0.4em\relax IEEE, 2005, pp. 75--80.

\bibitem{dellaert1999using}
F.~Dellaert, W.~Burgard, D.~Fox, and S.~Thrun, ``Using the condensation
  algorithm for robust, vision-based mobile robot localization,'' in
  \emph{Proceedings. 1999 IEEE computer society conference on computer vision
  and pattern recognition (Cat. No PR00149)}, vol.~2.\hskip 1em plus 0.5em
  minus 0.4em\relax IEEE, 1999, pp. 588--594.

\bibitem{5109666}
W.~Chung, S.~Kim, M.~Choi, J.~Choi, H.~Kim, C.-b. Moon, and J.-B. Song, ``Safe
  navigation of a mobile robot considering visibility of environment,''
  \emph{IEEE Transactions on Industrial Electronics}, vol.~56, no.~10, pp.
  3941--3950, 2009.

\bibitem{murray2000using}
D.~Murray and J.~J. Little, ``Using real-time stereo vision for mobile robot
  navigation,'' \emph{autonomous robots}, vol.~8, no.~2, pp. 161--171, 2000.

\bibitem{davison1999mobile}
A.~J. Davison, ``Mobile robot navigation using active vision,'' \emph{Advances
  in Scientific Philosophy Essays in Honour of}, 1999.

\bibitem{martinez2009bayesian}
R.~Martinez-Cantin, N.~De~Freitas, E.~Brochu, J.~Castellanos, and A.~Doucet,
  ``A bayesian exploration-exploitation approach for optimal online sensing and
  planning with a visually guided mobile robot,'' \emph{Autonomous Robots},
  vol.~27, no.~2, pp. 93--103, 2009.

\bibitem{4058690}
R.~Sim and J.~J. Little, ``Autonomous vision-based exploration and mapping
  using hybrid maps and rao-blackwellised particle filters,'' in \emph{2006
  IEEE/RSJ International Conference on Intelligent Robots and Systems}, 2006,
  pp. 2082--2089.

\bibitem{saulnier2020information}
K.~Saulnier, N.~Atanasov, G.~J. Pappas, and V.~Kumar, ``Information theoretic
  active exploration in signed distance fields,'' in \emph{2020 IEEE
  International Conference on Robotics and Automation (ICRA)}.\hskip 1em plus
  0.5em minus 0.4em\relax IEEE, 2020, pp. 4080--4085.

\bibitem{julian2014mutual}
B.~J. Julian, S.~Karaman, and D.~Rus, ``On mutual information-based control of
  range sensing robots for mapping applications,'' \emph{The International
  Journal of Robotics Research}, vol.~33, no.~10, pp. 1375--1392, 2014.

\bibitem{7139865}
B.~Charrow, S.~Liu, V.~Kumar, and N.~Michael, ``Information-theoretic mapping
  using cauchy-schwarz quadratic mutual information,'' in \emph{2015 IEEE
  International Conference on Robotics and Automation (ICRA)}, 2015, pp.
  4791--4798.

\bibitem{karlsson2005vslam}
N.~Karlsson, E.~Di~Bernardo, J.~Ostrowski, L.~Goncalves, P.~Pirjanian, and
  M.~E. Munich, ``The vslam algorithm for robust localization and mapping,'' in
  \emph{Proceedings of the 2005 IEEE international conference on robotics and
  automation}.\hskip 1em plus 0.5em minus 0.4em\relax IEEE, 2005, pp. 24--29.

\bibitem{coffin2022multi}
H.~Coffin, I.~Abraham, G.~Sartoretti, T.~Dillstrom, and H.~Choset,
  ``Multi-agent dynamic ergodic search with low-information sensors,'' in
  \emph{2022 International Conference on Robotics and Automation (ICRA)}.\hskip
  1em plus 0.5em minus 0.4em\relax IEEE, 2022, pp. 11\,480--11\,486.

\bibitem{9312988}
I.~Abraham, A.~Prabhakar, and T.~D. Murphey, ``An ergodic measure for active
  learning from equilibrium,'' \emph{IEEE Transactions on Automation Science
  and Engineering}, vol.~18, no.~3, pp. 917--931, 2021.

\bibitem{mathew2011metrics}
G.~Mathew and I.~Mezi{\'c}, ``Metrics for ergodicity and design of ergodic
  dynamics for multi-agent systems,'' \emph{Physica D: Nonlinear Phenomena},
  vol. 240, no. 4-5, pp. 432--442, 2011.

\bibitem{de2016ergodic}
G.~De~La~Torre, K.~Fla{\ss}kamp, A.~Prabhakar, and T.~D. Murphey, ``Ergodic
  exploration with stochastic sensor dynamics,'' in \emph{2016 American Control
  Conference (ACC)}.\hskip 1em plus 0.5em minus 0.4em\relax IEEE, 2016, pp.
  2971--2976.

\bibitem{kullback1951information}
S.~Kullback and R.~A. Leibler, ``On information and sufficiency,'' \emph{The
  annals of mathematical statistics}, vol.~22, no.~1, pp. 79--86, 1951.

\bibitem{rubner2000earth}
Y.~Rubner, C.~Tomasi, and L.~J. Guibas, ``The earth mover's distance as a
  metric for image retrieval,'' \emph{International journal of computer
  vision}, vol.~40, no.~2, p.~99, 2000.

\bibitem{boyd2004convex}
S.~Boyd, S.~P. Boyd, and L.~Vandenberghe, \emph{Convex optimization}.\hskip 1em
  plus 0.5em minus 0.4em\relax Cambridge university press, 2004.

\bibitem{lukvsan2004interior}
L.~Luk{\v{s}}an, C.~Matonoha, and J.~Vl{\v{c}}ek, ``Interior-point method for
  non-linear non-convex optimization,'' \emph{Numerical linear algebra with
  applications}, vol.~11, no. 5-6, pp. 431--453, 2004.

\bibitem{elfes1989using}
A.~Elfes, ``Using occupancy grids for mobile robot perception and navigation,''
  \emph{Computer}, vol.~22, no.~6, pp. 46--57, 1989.

\bibitem{coumans2021}
E.~Coumans and Y.~Bai, ``Pybullet, a python module for physics simulation for
  games, robotics and machine learning,'' \url{http://pybullet.org},
  2016--2021.

\bibitem{rover}
N.~Ulmasov, J.~Siri, J.~Wallace, M.~Mei, and C.~Buscaron, ``Aws robomaker
  sample application open source rover,''
  \url{https://github.com/aws-samples/aws-robomaker-sample-application-open-source-rover},
  2020.

\bibitem{MANGOLD20171}
\BIBentryALTinterwordspacing
N.~Mangold, M.~Schmidt, M.~Fisk, O.~Forni, S.~McLennan, D.~Ming, V.~Sautter,
  D.~Sumner, A.~Williams, S.~Clegg, A.~Cousin, O.~Gasnault, R.~Gellert,
  J.~Grotzinger, and R.~Wiens, ``Classification scheme for sedimentary and
  igneous rocks in gale crater, mars,'' \emph{Icarus}, vol. 284, pp. 1--17,
  2017. [Online]. Available:
  \url{https://www.sciencedirect.com/science/article/pii/S0019103516302949}
\BIBentrySTDinterwordspacing

\bibitem{tensorflow2015-whitepaper}
\BIBentryALTinterwordspacing
M.~Abadi, A.~Agarwal, P.~Barham, E.~Brevdo, Z.~Chen, C.~Citro, G.~S. Corrado,
  A.~Davis, J.~Dean, M.~Devin, S.~Ghemawat, I.~Goodfellow, A.~Harp, G.~Irving,
  M.~Isard, Y.~Jia, R.~Jozefowicz, L.~Kaiser, M.~Kudlur, J.~Levenberg,
  D.~Man\'{e}, R.~Monga, S.~Moore, D.~Murray, C.~Olah, M.~Schuster, J.~Shlens,
  B.~Steiner, I.~Sutskever, K.~Talwar, P.~Tucker, V.~Vanhoucke, V.~Vasudevan,
  F.~Vi\'{e}gas, O.~Vinyals, P.~Warden, M.~Wattenberg, M.~Wicke, Y.~Yu, and
  X.~Zheng, ``{TensorFlow}: Large-scale machine learning on heterogeneous
  systems,'' 2015, software available from tensorflow.org. [Online]. Available:
  \url{https://www.tensorflow.org/}
\BIBentrySTDinterwordspacing

\end{thebibliography}
\balance

\end{document}